\documentclass[a4paper,fleqn]{cas-sc}

\usepackage{algorithm,algorithmic}
\usepackage[section]{placeins}

\newcommand{\tabincell}[2]{
\begin{tabular}{@{}#1@{}}#2\end{tabular}
}

\usepackage[numbers]{natbib} 

\def\tsc#1{\csdef{#1}{\textsc{\lowercase{#1}}\xspace}}
\tsc{WGM}
\tsc{QE}
\tsc{EP}
\tsc{PMS}
\tsc{BEC}
\tsc{DE}

\begin{document}
\let\WriteBookmarks\relax
\def\floatpagepagefraction{1}
\def\textpagefraction{.001}
\shorttitle{Clustering through Sequence Analysis}
\shortauthors{ShiGB et~al.}

\title [mode = title]{Clustering through Feature Space Sequence Discovery and Analysis}                     
\tnotemark[1]


\author[1]{Shi Guobin}[type=editor,
                        auid=000,bioid=1,
                        orcid=0000-0001-8675-0184]
\address[1]{Department of Electronic and Information Engineering 1310, Lan Zhou Jiao Tong University, LanZhou 730070,GanSu Province, China}


\cormark[1]
\ead{tubing86@126.com}

\tnotetext[1]{This research did not receive any specific grant from funding agencies in the public, commercial, or not-for-profit sectors.}
\cortext[cor1]{Corresponding author}

\nonumnote{In this work we proposed a new algorithm which can automatically find cluster in high-dimensional data feature space in a simple way.
  }

\begin{abstract}
Identifying high-dimensional data patterns without a priori knowledge is an important task of data science.  This paper proposes a simple and efficient noparametric algorithm: Data Convert to Sequence Analysis, DCSA, which dynamically explore each point in the feature space without repetition, and a Directed Hamilton Path will be found.  Based on the change point analysis theory, The sequence corresponding to the path is cut into several fragments to achieve clustering.  The experiments on real-world datasets from different fields with dimensions ranging from 4 to 20531 confirm that the method in this work is robust and has visual interpretability in result analysis.
\end{abstract}


\maketitle

\section{Introduction}

\begin{figure}[pos=H]

\centering
\includegraphics[scale=0.6]{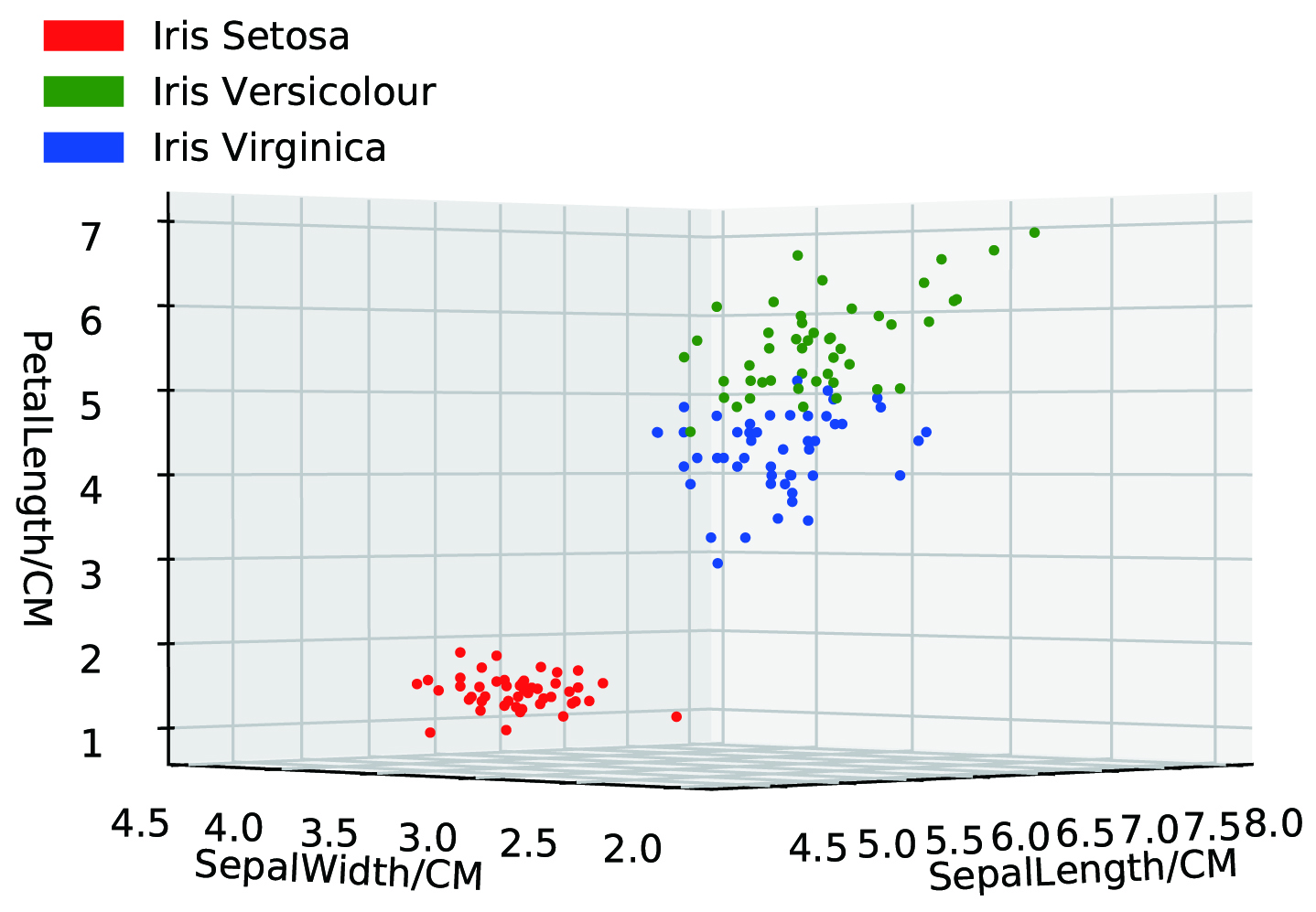}
\caption{3D diagram of the iris data set}
\label{fig:1}
\end{figure}

Data points are embedded in the feature space constructed by their attributes.  If the data has a natural classification, its spatial structure may have several forms of distinguishability.  Figure \ref{fig:1} is a 3D image generated by the Iris data set\citep{dua2017uci} with three attributes.  It is noticed that the red class is obviously distinguished from other two classes.  We assume that the information is compressed in all dimensions, if all four dimensions can be used and displayed, the distinguishability of the blue and green categories in Figure \ref{fig:1} will be more obvious.

In this work, we designed a Data Convert To Sequence Analysis Algorithm (DCSA) to prove this hypothesis.  The DCSA  is capable to explore the space constructed by full-dimensional of data, looking for the distinguishable clues, and then discover clusters.  In this paper, the terminology “multi-dimension” refers to space dimension with more than 3, while “high-dimension” referes to space dimension frome dozens to ten thousands.

The DCSA takes the euclidean distance of data points to generate a directed Hamilton Path\citep{angluin1979fast} to connect all points.  The distance between every two points on the path forms a sequence which is further divided into fragments using sequence analysis technology, and each segment is considered as a class.

The traditional clustering method has successfully solved the clustering problem of low-dimensional data\citep{bouveyron2014model}.However, the clustering of the multi-dimension, especially the high-dimensional data is more complicated because of the following reasons: 

1.  The probability of cluster structure in the full attributes feature space is lower; 

2.  The data distribution in the high-dimensional space is more sparse, which has an effect on the discovery of cluster structure.

High-dimensional data pattern recognition several methods proposed in the literature: such as dimensionality reduction\citep{sanguinetti2008dimensionality}, regularization-based technology\citep{bickel2008covariance}, parsimonious modeling \citep{celeux1995gaussian}, subspace clustering\citep{parsons2004subspace} and variable selection-based clustering\citep{raftery2006variable}.  The dimensionality reduction method is classified as follows:

\scriptsize{\[{{\begin{array}{l}
Dimensional\\
Reduction
 \end{array}}}\left\{ {\begin{array}{l}
{{{Linear}}\left\{{\begin{array}{l}
{{PCA}}\\
\begin{array}{l}
{{LDA}}\\
{{Spectral\ Clusetring}}
\end{array}
\end{array}} \right.}\\
{{{\begin{array}{l}
Non\\
Linear
 \end{array}}}\left\{ \begin{array}{ll}
{{\begin{array}{l}
Keep\ local\\
attributes
 \end{array}}}\left\{ {\begin{array}{l}
{{{LLE}}}\\
{{{Laplacian\ Eigenmaps}}}\\
{{{Tangent\ space}}\left\{ {\begin{array}{l}
{{{Hessian\ LLE}}}\\
{{{LTSA}}}
\end{array}} \right.}
\end{array}} \right.\\
{{\begin{array}{l}
Keep\ global\\
attributes
 \end{array}}}\left\{ {\begin{array}{l}
{{{\begin{array}{l}
based\ on\\
distance
 \end{array}}}\left\{ {\begin{array}{l}
{{{L2\ distance:MDS}}}\\
{{{Geodesic\ distance:ISO\ map}}}\\
{{{Dispersion\ distance:diffusion\ maps}}}
\end{array}} \right.}\\
{{{based\ on\ kernel}}}\\
{{{based\ on\ neural\ network}}}
\end{array}} \right.
\end{array} \right.}
\end{array}} \right.\]}

\normalsize
DCSA is a dimensionality reduction algorithm based on distance preserving global attributes.

\section{Data Convert To Sequence}
\subsection{2.1 Sequence discovery}
Feature space of the dataset exists directed graph\citep{harary1965structural} ${{G}_{i}}=(V, {{E}_{i}})$ , which is a $(p, q)$ graph, where $p$ is the number of sample points, and $q$ is the edge that connect all points. If there exists a graph $G$ in ${{G}_{i}}$which represents a path and each vertex for G was visited exactly once, and $|q|=|p|-1$, then $G$ identifies a Hamiltonian path.

The DCSA algorithm for finding the Hamiltonian path is shown in the Algorithm1:
\begin{algorithm}[H] 
\caption{ Data Convert to Sequence Analysis algorithm} 
\label{alg:Framwork} 
\begin{algorithmic}[1] 
\REQUIRE Data set matrix ${{\mathbf{X}}_{p\times k}}$\\ 
\ENSURE Adjacency matrix $\mathbf{N}$ , Dictionary $Dict$\\ 
\STATE Create empty matrix ${{\mathbf{N}}_{p\times p}}=\mathbf{0}$, empty dictionary class variable $Dict$; 
\label{ code:fram:CREATE }
\STATE Use the data matrix $\mathbf{X}$ to calculate the Gram matrix $\mathbf{G}$; 
\label{code:fram:caculateG}
\STATE Calculate Euclidean distance matrix $\mathbf{D}$ by $\mathbf{G}$; 
\label{code:fram:caculateD}
\STATE Select a row of $\mathbf{D}$ as a starting point by calculation, denoted as $start$; 
\label{code:fram:selectR}
\STATE Find the smallest ${{q}_{next}}$ in row $start$ of $\mathbf{D}$, set ${{\mathbf{N}}_{start.next}}={{q}_{i}}$,  Append record $\{{{x}_{start}}\to {{x}_{next}}:{{q}_{i}}\}$ to $Dict$.; 
\label{code:fram:FindMin}
\STATE Set ${{\mathbf{D}}_{column=next}}=\infty $, $start=\rm next$.  Loop execution 4-5 till $\mathbf{D}=\infty $(each item in $\mathbf{D}$ eaquals to ``$\inf $'' which means $\infty $); 
\label{code:fram:result}
\RETURN $\mathbf{N}$ $Dict$; 
\end{algorithmic}
\end{algorithm}

GramMatrix\citep{horn2012matrix}in Step\ref{code:fram:caculateG} of Algorithm 1 is calculated as:
\begin{equation}
 \mathbf{G}=\mathbf{X}{{\mathbf{X}}^{\top }}
\end{equation}

Its time complexity is $O({{p}^{2}}k)$.  In fact, it is not necessary to calculate the Gram matrix, just randomly choose a starting point, and the rest is determined by dynamic planning. Otherwise, when the sample over 100,000, the size of G will exceed 70GB because of Gram matrix space complexity is $O({{p}^{2}})$. We calculate G to ensure that the experiment can be reproduced. If a starting point is randomly selected, the time complexity of the algorithm will be reduced to $O({{p}^{2}}k/2)$, and space complexity will sharply reduced to $O(1)$ .

The Euclidean distance matrix D in Setp\ref{code:fram:caculateD} is calculated as\citep{parhizkar2013euclidean}:
\begin{equation}
\mathbf{D}=diag(\mathbf{G}).{{\mathbf{1}}^{\top }}-2\mathbf{G}+\mathbf{1}.diag{{(\mathbf{G})}^{\top }}
\end{equation}

The method to find the starting point in Setp\ref{code:fram:selectR} is as follows:

The row index of the distance matrix $\mathbf{D}$ represents the ordinal number of the data element, and the row sums of $\mathbf{D}$ represents the total distance from the element to the others, including itself. The row index which has the smallest row-sum can be regarded as the center of a highest density area.

The order of the $Dict$ value, is the right sequence we are looking for, here marked as ${{\alpha }_{n}}(n\in \{1, 2, 3...p\})$.

The time complexity of Step 3-5 is $O({{p}^{2}}/2)$, and the total time complexity of steps 0-6 is $O\left( {{p}^{2}}(k+0.5) \right)$.

DCSA is a dynamic programming algorithm, and its model is as follows:
\begin{equation}
{{h}_{1p}}({{x}_{1}})=\{{{u}_{1}}({{x}_{1}}), {{u}_{2}}({{x}_{2}}), \cdots , {{u}_{p}}({{x}_{p}})\}
\end{equation}

where ${{h}_{1p}}$ represents the entire decision-making process from the starting point 1 to the end point $p$.  ${{u}_{1}}({{x}_{1}})$ represents the set of decisions in the state ${{x}_{1}}$.In this paper:
\[\begin{array}{l}
{u_1}({x_1}){\rm{ = }}\mathop {{\rm{argmin}}}\limits_{{x_j} \in D} \sum\limits_k {\sqrt {{{(x_1^k - x_j^k)}^2}} }  = {x_2}{\rm{  s}}{\rm{.t}}{\rm{.  }}{(x_1^k - x_j^k)^2} \ne 0\\
 \cdots \\
{u_i}({x_i}){\rm{ = }}\mathop {{\rm{argmin}}}\limits_{{x_j} \in D} \sum\limits_k {\sqrt {{{(x_i^k - x_j^k)}^2}} }  = {x_{i + 1}}{\rm{  s}}{\rm{.t}}{\rm{.  }}{(x_i^k - x_j^k)^2} \ne 0\\
 \cdots \\
{u_{p - 1}}({x_{p - 1}}){\rm{ = }}\mathop {{\rm{argmin}}}\limits_{{x_j}\in D} \sum\limits_k {\sqrt {{{(x_{p - 1}^k - x_j^k)}^2}} }  = {x_p} = {x_{end}}
\end{array}\]

where ${{x}_{i}}\text{=(}x_{i}^{1}, x_{i}^{2}, \cdots , x_{i}^{k}{{)}^{\top }}$, $D=\{x_{1}^{k}, x_{2}^{k}, \cdots , x_{p}^{k}\}$

The process of the algorithm for finding the Hamiltonian path is a Markov decision process\citep{bellman1957markovian, markov1906extension, seneta2006markov}.  The current state is only affected by its previous state and has nothing to do with the other historical states.  The formal expression is as follows:
\begin{equation}
P({{x}_{t}}|{{x}_{t-1}}, {{u}_{t-1}}, {{x}_{t-2}}, {{u}_{t-2}}\cdots , {{x}_{1}}, {{u}_{1}})=P({{x}_{t}}|{{x}_{t-1}}, {{u}_{t-1}})
\end{equation}

Generally speaking, in the data convert to sequence stage, we assume that there is a discovery factor (DF), which will explore $p$ points in an $k$ dimensional space.  The starting time for exploration is $t_{start}^{1}$, and the starting point ${{p}_{start}}$ is obtained by calculation .DF starts from the current point to the next one with the closest Euclidean distance, and does not visit a point twice.  After $t_{end}^{p}-t_{start}^{1}$ time, it will visit all points and arrive ${{p}_{end}}$, leaving an exploration path composed of $p-1$ distances.

The experiment for DCSA shows that the majority of the same class point is concentrated in the same interval of the sequence.

\begin{figure}[pos=H]
\centering
\includegraphics[scale=0.33]{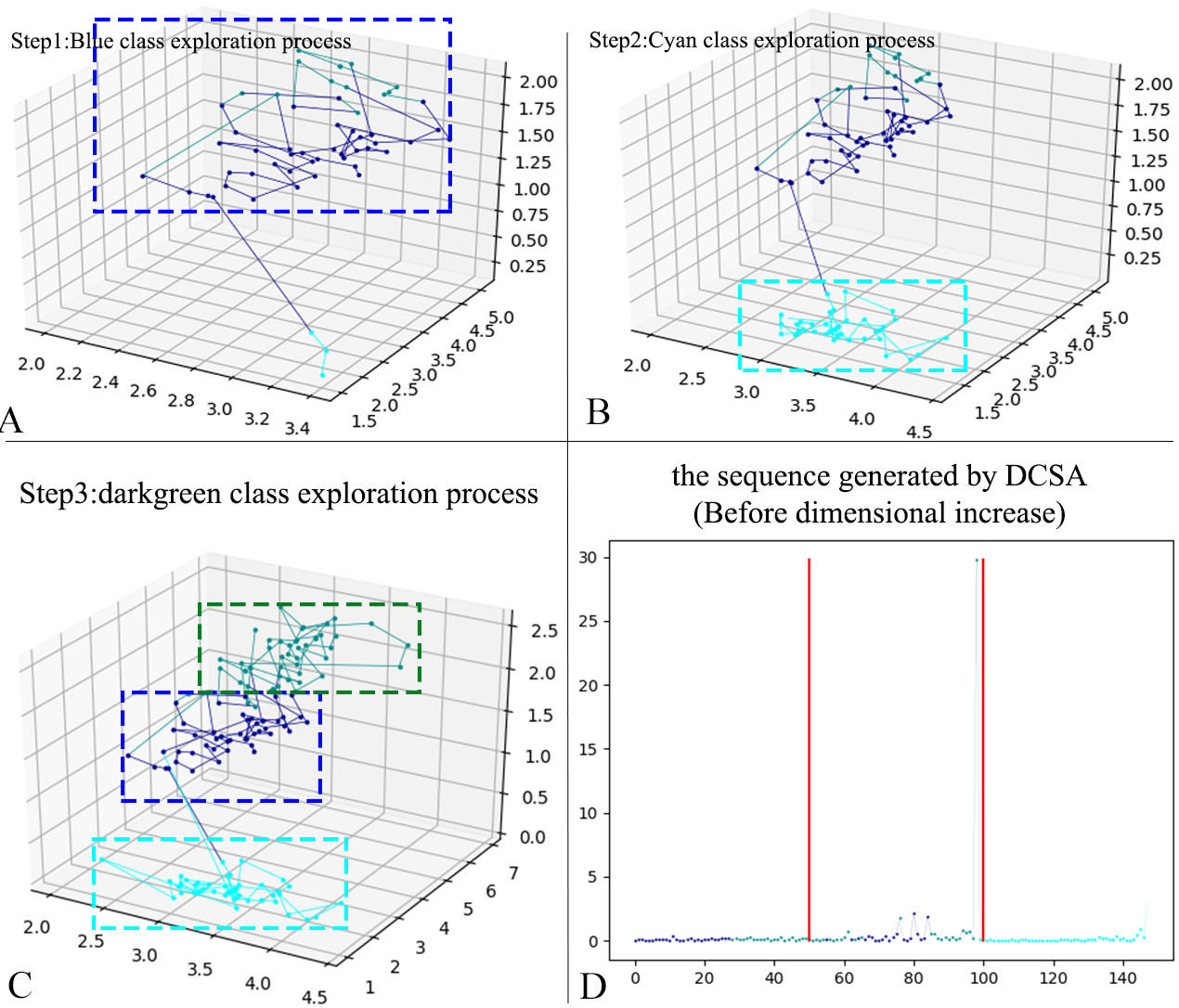}
\caption{DCSA exploration process on the Iris data set(Before demension incresase optimization)}
\label{fig:3}
\end{figure}

The actual exploration process of DCSA on the original iris data set is shown in Figure \ref{fig:3}A to C.  The purple, cyan, and green represent iris-setosa, iris-versicolour, and iris-virginica, respectively.

It is shown that the DCSA exploration is carried out in a priority class, especially the cyan with a large distance from other classes has 100\% exploration continuity. The purple class and the green class are overlap in space, leading to alternate exploration processes. We speculate that the DCSA application will have better results by mapping data to higher dimensions through the kernel method\citep{hofmann2008kernel}.Figure \ref{fig:3}D shows the coloring result of the DCSA generated sequence.

\subsection{Data standardization}
The DCSA is sensitive to the differences in the value range of data attributes, and such data needs to be standardized pre-processing\citep{alasadi2017review}.
\begin{figure}[pos=H]
\centering
\includegraphics[scale=0.56]{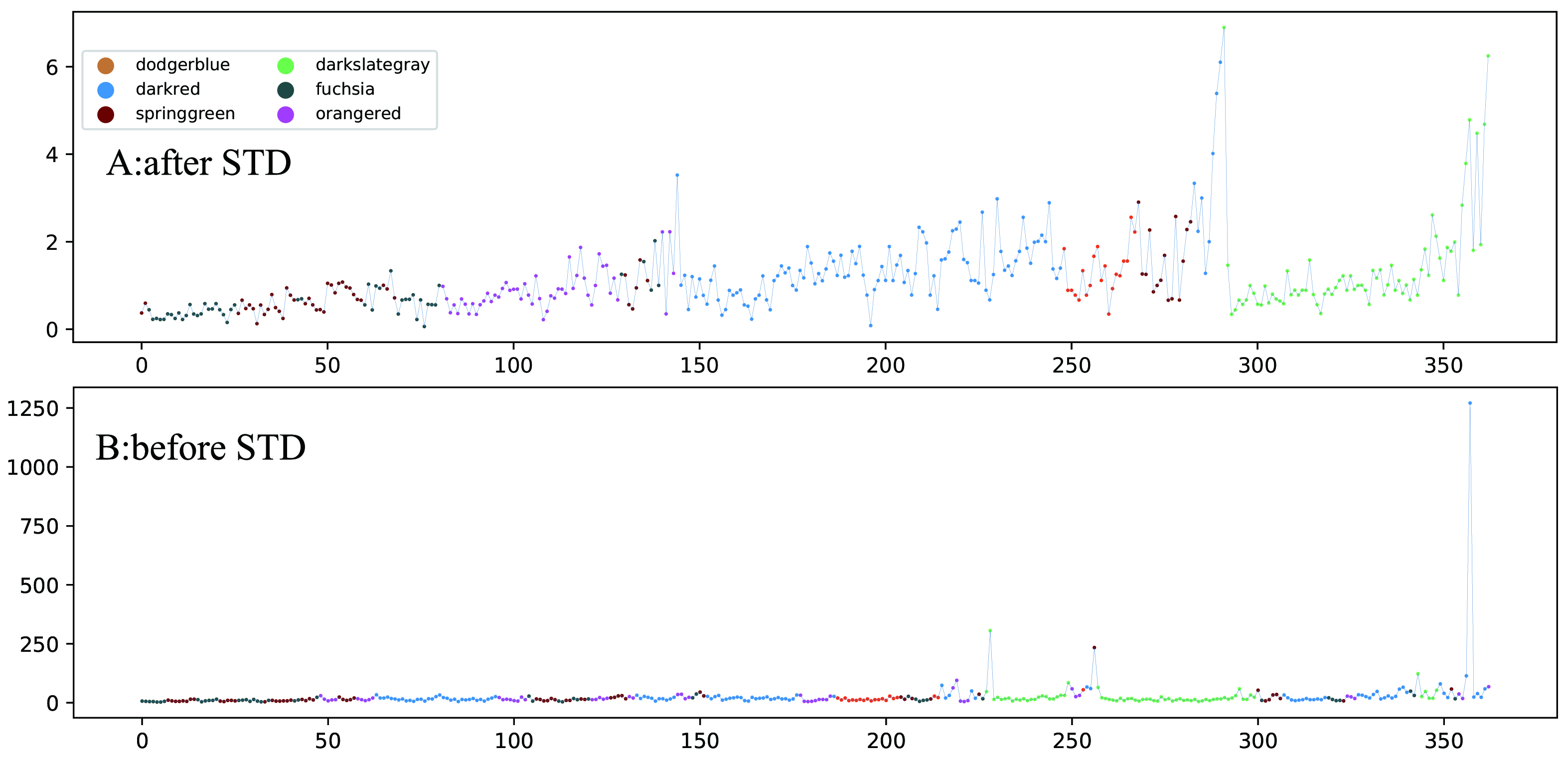}
\caption{The impact of standardization on DCSA}
\label{fig:4}
\end{figure}

As shown in Figure\ref{fig:4}, the obviously enhancement for the regularity can be seen after standardization. With coloring the sequence by class, it is clearly showed that the effectiveness of the algorithm DCSA after standardization.
\subsection{Low-dimensional data application DCSA}
\begin{figure}[pos=H]
\centering
\includegraphics[scale=0.45]{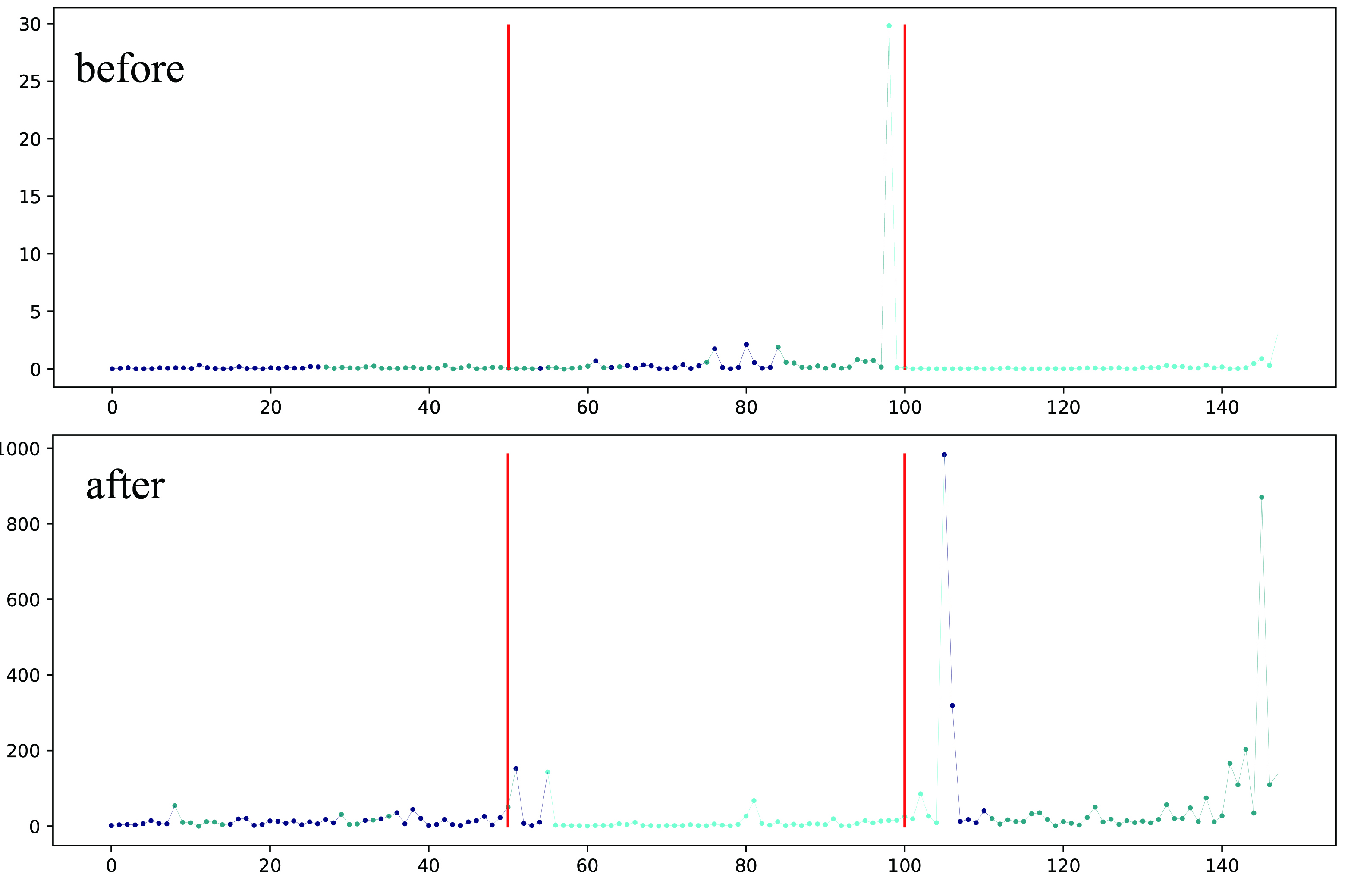}
\caption{Comparison before and after dimension increase}
\label{fig:5}
\end{figure}

It is noticed in Figure \ref{fig:3}D that the DCSA application is not ideal, resulting from that the Iris has only four attributes. Therefore, we use polynomial kernel\citep{chang2010training} to increase the dimension for the further application of the DCSA. The result is plotted in Figure \ref{fig:5}, of which the red vertical line is the known dividing point.

After increasing the dimension, the sequence has obvious jumps between different classes. It is demonstrated that the continuity of similar data points in the same interval is enhanced and the effect of the DCSA is also improved.

\section{Sequence Analysis}
\subsection{Theoretical basis}
\begin{figure}[pos=H]
\centering
\includegraphics[scale=0.5]{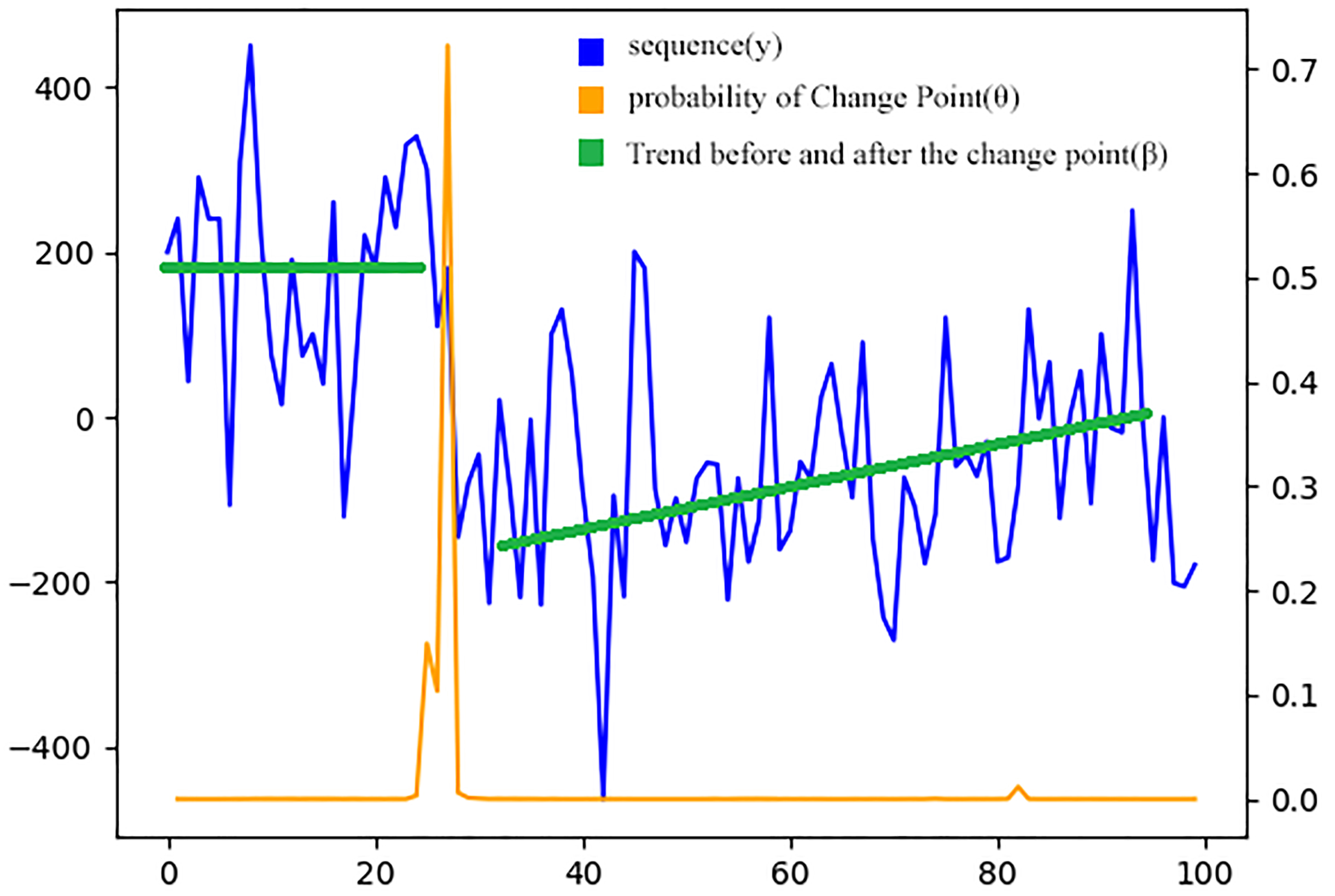}
\caption{Schematic diagram of the general method of change point analysis}
\label{fig:6}
\end{figure}

Li A. G. et al.\citep{0On} supported that sequence truncation, segmentation, clustering, etc. could be unified into Sequence Change Point Analysis (SCPA) problem, First we divide the sequence into 1-K segments,  and each segment obeys one of the orthogonal transformation model set M (M can be linear polynomial model, wavelet transform Fourier transform model etc). The problem of sequence change point is shown in Figure \ref{fig:6}, where the sequence before and after the change point is represented by a linear model (green line). This is also an optimal segmentation problem.

\subsection{Mathematical modeling}
The optimal segmentation problem is modeled as follows.
The segment modeling for original sequence $x={{x}_{1}}, {{x}_{2}}, ..., {{x}_{N}}$ is as follows:
\[x=\left\{ \begin{matrix}
   f(t, {{w}_{1}})+{{e}_{1}}(t), \text{   }1\le t<{{\alpha }_{1}}  \\
   ...   \\
   f(t, {{w}_{N}})+{{e}_{k}}(t), \text{   }{{\alpha }_{k-1}}\le t<{{\alpha }_{k}}  \\
\end{matrix} \right.\]
Where $f$ is the function representation of each segment, ${{e}_{i}}(t)\tilde{\ }N(0, {{\sigma }^{2}})$ is noise, and ${{\alpha }_{i}}$ is the segment point.
The objective function is to minimize the mean-square error(MSE) distance between segment function $f$ and those points in this segment:
\[L=\sum\limits_{i=1}^{k}{{{l}_{i}}}=\sum\limits_{i=1}^{k}{\sum\limits_{j=0}^{{{m}_{i}}}{{{({{x}_{{{\alpha }_{i-1}}+j}}-{{f}_{i}}({{\alpha }_{i-1}}+j, {{w}_{i}}))}^{2}}}}\]
${{m}_{i}}={{\alpha }_{i}}-{{\alpha }_{i-1}}$ is the number of samples in segment $i$.
If the number of segments is not limited, and every two points can be identified as a segment, the objective function value is 0.  Therefore the following constraints are required:
\[J = \left\{ {\begin{array}{*{20}{c}}
\begin{array}{l}
\frac{L}{{\sum\limits_{i = 1}^N {{{({x_i} - \bar x)}^2}} }} + \frac{k}{N}, {\rm{     }}\sum\limits_{i = 1}^N {{{({x_i} - \bar x)}^2}} {\rm{ > 0}}\\
 \cdots 
\end{array}\\
{\frac{k}{N}, {\rm{      }}\sum\limits_{i = 1}^N {{{({x_i} - \bar x)}^2} = 0} }
\end{array}} \right.\]

The above is the basic method for sequence change point discovery.  The application based on the above theory are further detailed in the following section.
\subsection{Related algorithms}
In present papper, we select four sequence analysis algorithms:  BCD,CUSUM,ARIMA,JNBD. The BCD algorithm is a newer algorithm, which will be detailedly described here, and only a brief introduction of other three methods will be given here, due to they are all well-known algorithms.
\subsubsection{BCD algorithm}
Xiang et al.\citep{xuan2007bayesian, fearnhead2006exact, adams2007bayesian, xuan2007modeling} used Bayesian method to calculate the probability of change points (Bayesian Changepoint Detection, BCD). Similar to the theoretical basis, the problem is transformed into a mixed linear model containing change points:
\begin{equation}
y(t)={{\beta }_{0}}H_{-}^{\theta }+{{\beta }_{1}}H_{+}^{\theta }+\xi (t)
\label{model:5}
\end{equation}

Where ${{\beta }_{{}}}$ is the linear fitting coefficient before and after the discontinuity point, which is used to describe the changing trend of the sequence before and after the change point; $\xi \tilde{\ }N(0, {{\sigma }^{2}}\Omega )$ is the noise,  $\mathbf{\Omega }$ is the variance matrix; $H$ is the step function, which is used to distinguish between before and after the change point:
\[H_ - ^\theta  = \left\{ \begin{array}{l}
1\qquad{\rm{   if\  t <  = }}\theta \\
0\qquad{\rm{  else}}
\end{array} \right.\]
\[H_ + ^\theta  = \left\{ \begin{array}{l}
1\qquad{\rm{   if\  t >  = }}\theta \\
0\qquad{\rm{  else}}
\end{array} \right.\]

The above model(\ref{model:5}) can be rewritten as:
\begin{equation}
y(t)=\mathbf{\beta H}+\xi \iff\ y\sim{}N(\mathbf{H}\overset{\wedge }{\mathop{\mathbf{\beta }}}\, , {{\sigma }^{2}}\mathbf{\Omega })
\end{equation}

Applying Bayesian formula, we have
\begin{equation}
p(\theta |y)=\frac{p(y|\theta )p(\theta )}{p(y)}
\end{equation}

when $p(y)$ is known as an observation, and the equivalent equation as:
\begin{equation}
p(\theta |y)\sim p(y|\theta )p(\theta )=p(y\theta )
\end{equation}

We take the hidden variable $\theta $ (change point) into consideration, and the corresponding likelihood equation of Formula (6) is:
\begin{equation}
L(\theta , \mathbf{\beta }, \sigma , \mathbf{\Omega }|y)=\frac{1}{{{(2\pi {{\sigma }^{2}})}^{\frac{n}{2}}}\sqrt{|\mathbf{\Omega }|}}{{e}^{-\frac{1}{2{{\sigma }^{2}}}{{(y-\mathbf{H\beta })}^{\top }}\mathbf{\Omega }(y-\mathbf{H\beta })}}
\end{equation}

This is the likelihood function form of Formula 6, which is corresponded to Bayesian function Formula 8.

Parameter ${{\beta }_{{}}}$ is unknown, there must be a ${{\beta }_{{}}}^{*}$ that maximizes the likelihood equation:
\begin{equation}
L(\theta , \beta , \sigma , {\bf{\Omega }}) = \frac{1}{{{{(2\pi {\sigma ^2})}^{\frac{n}{2}}}\sqrt {|{\bf{\Omega }}|} }}{e^{ - \frac{{{{\cal R}^2}}}{{2{\sigma ^2}}}}}{e^{\frac{1}{{2{\sigma ^2}}}{{(\beta  - {\beta ^*})}^ \top }\Xi (\beta  - {\beta ^*})}}
\end{equation}

where $\Xi  = {{\bf{H}}^ \top }{\bf{\Omega H}}$, ${{\mathcal{R}}^{2}}$ is the residual.

Assuming that all parameters are independent, then:
\[{{p}_{y}}(\theta , \beta , \sigma )\sim{\ }L(\theta , \beta , \sigma )\]

Integrate $\beta $ and $\sigma $ to get $p(\theta )$.  Those points which is corresponded to the largest
 $p(\theta )$ is the division point of the sequence.

The experiment in this paper used the BCD implementation library from python: bayesian\_changepoint\_detection.
\subsubsection{CUSUM algorithm}
Cumulative Sum Control Chart\citep{hawkins1981cusum, hawkins2012cumulative} (CUSUM) is a sequential analysis method that accumulates small deviations in the process to achieve the effect of amplification, leading to an improvement on the sensitivity to small deviations during the detection process. When CUSUM detects the cumulative deviation  is significantly higher or lower than the average level under normal and stable operating conditions, it means that the system has changed. The self-edited algorithm CUSUM-A and the Python package detecta\.detect\_cusum(CUSUM-B) are applied in this work.
\subsubsection{ARIMA algorithm}
Autoregressive Integrated Moving Average model\citep{asteriou2011arima}(ARIMA) has three parameters (p, d, q), p is the number of autoregressive items; MA is "moving average", q is the number of moving average items, and d is the number of items that delive a stationary sequence.  The ARIMA model can be expressed as:
\[(1-\sum\limits_{i=1}^{p}{{{\phi }_{i}}{{L}^{i}}}){{(1-L)}^{d}}{{X}_{t}}=(1+\sum\limits_{i=1}^{q}{{{\theta }_{i}}{{L}^{i}}}){{\varepsilon }_{t}}\]

$L$ is Lag operator, $d\in \mathbb{Z}, d>0$.

The experiment in this paper used the ARIMA implementation library from python: Change\_Finder.
\subsubsection{JNBD algorithm}
Finally we applied the Jenks natural breaks points detection algorithm\citep{jenks1967data} (JNBD).  This method is almost completely consistent with the method introduced in the theoretical basis.  Initializing random breakpoints, and then iteratively calculating different breakpoint position, The truncation principle is to maximize variance between groups and minimize variance within groups.  Corresponding library in Python is jenkspy.

\section{Experiments}
The experimental are performed in the following computational configuration: IntelXeonSilver\-4110 CPU 2.10GHz, 48GB DDR4 memory and Windows10-64bit enterprise operating system with Python 3.7.
\subsection{Application of artificial data sets}
\subsubsection{Overall performance test}
\begin{figure}[pos=H]
\centering
\includegraphics[width=1\textwidth]{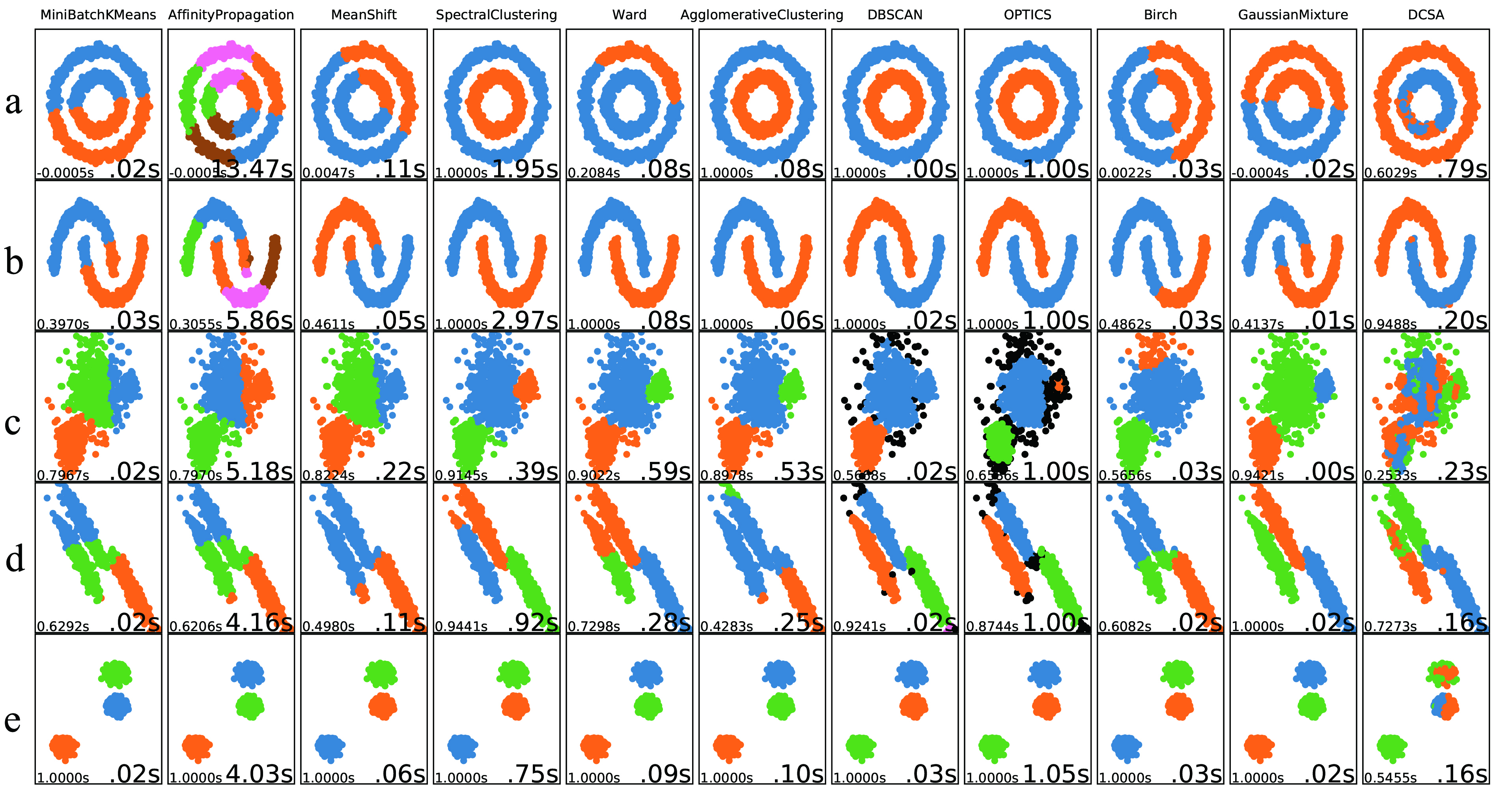}
\caption{Comparison of clustering algorithms}
\label{fig:7insert}
\end{figure}

Figure \ref{fig:7insert} is shows an overall comparison between DCSA and some classic clustering algorithms on the low-dimensional generated dataset(a:'noisy\_circles', b:'noisy\_moons', c:'blobs', d:'aniso', e:'varied').The AMI score is in the lower left corner of each chart, and the elapsed time is in the lower right foot.  It is obvious that K-means algorithm has the best operating efficiency.
The DCSA algorithm in Figure \ref{fig:7insert} performs well on the dataset a, b, and d, all of which are with non-cluster structures. On the other hand, in the dataset c and e, especially the dataset c, the result shows that DCSA is invalid for low-dimensional data with fuzzy boundaries.

Table 1 shows the Adjusted Mutual Info(AMI) score of each algorithm.  The algorithm numbers are:  A1:MiniBatchK\\Means, A2:AffinityPropagation, A3:MeanShift, A4:SpectralClustering, A5:Ward, A6:AgglomerativeClustering, A7:D\\BSCAN, A8:OPTICS, A9:Birch, A10:GaussianMixture, A11:DCSA(with CUSUM-B).
\begin{table}[pos=H].As shown in Table 1, the AMI score of DCSA is rank 7/11, and the best combined performer across the five datasets was SpectralClustering.
\centering\small
\resizebox{\textwidth}{21mm}{
\begin{tabular}{llllllllllll}
\multicolumn{12}{l}{\small{\textbf{Table 1}}}\\
\multicolumn{12}{l}{\small{Scores of 11 algorithms on 5 generated datasets}}\\\specialrule{0.05em}{3pt}{3pt}
$Datasets$&A1	&A2	&A3	&A4	&A5 &A6 &A7 &A8 &A9 &A10 &A11\\\specialrule{0.05em}{2pt}{2pt}
$a:noisy\_circles$&0&0&0.005&1&0.208&1&1&1&0.002&0&0.603\\[7pt]
$b:noisy\_moons$&0.397&0.305&0.461&1&1&1&1&1&0.486&0.413&0.949\\[7pt]
$c:blobs$&0.796&0.797&0.822&0.914&0.902&0.898&0.566&0.656&0.565&0.9421&0.259\\[7pt]
$d:aniso$&0.629&0.621&0.498&0.944&0.73&0.4283&0.9241&0.874&0.608&1&0.727\\[7pt]
$e:varied$&1&1&1&1&1&1&1&1&1&1&0.546\\\specialrule{0.05em}{2pt}{2pt}
$Mean:Rank$&0.564:8&0.545:10&0.557:9&0.972:1&0.768:5&0.865:4&0.898:3&0.906:2&0.532:11&0.671:6&0.617:7\\\specialrule{0.05em}{2pt}{2pt}
\end{tabular}}
\end{table}

\subsubsection{High dimensional test}
\begin{figure}[pos=H]
\centering
\includegraphics[scale=0.35]{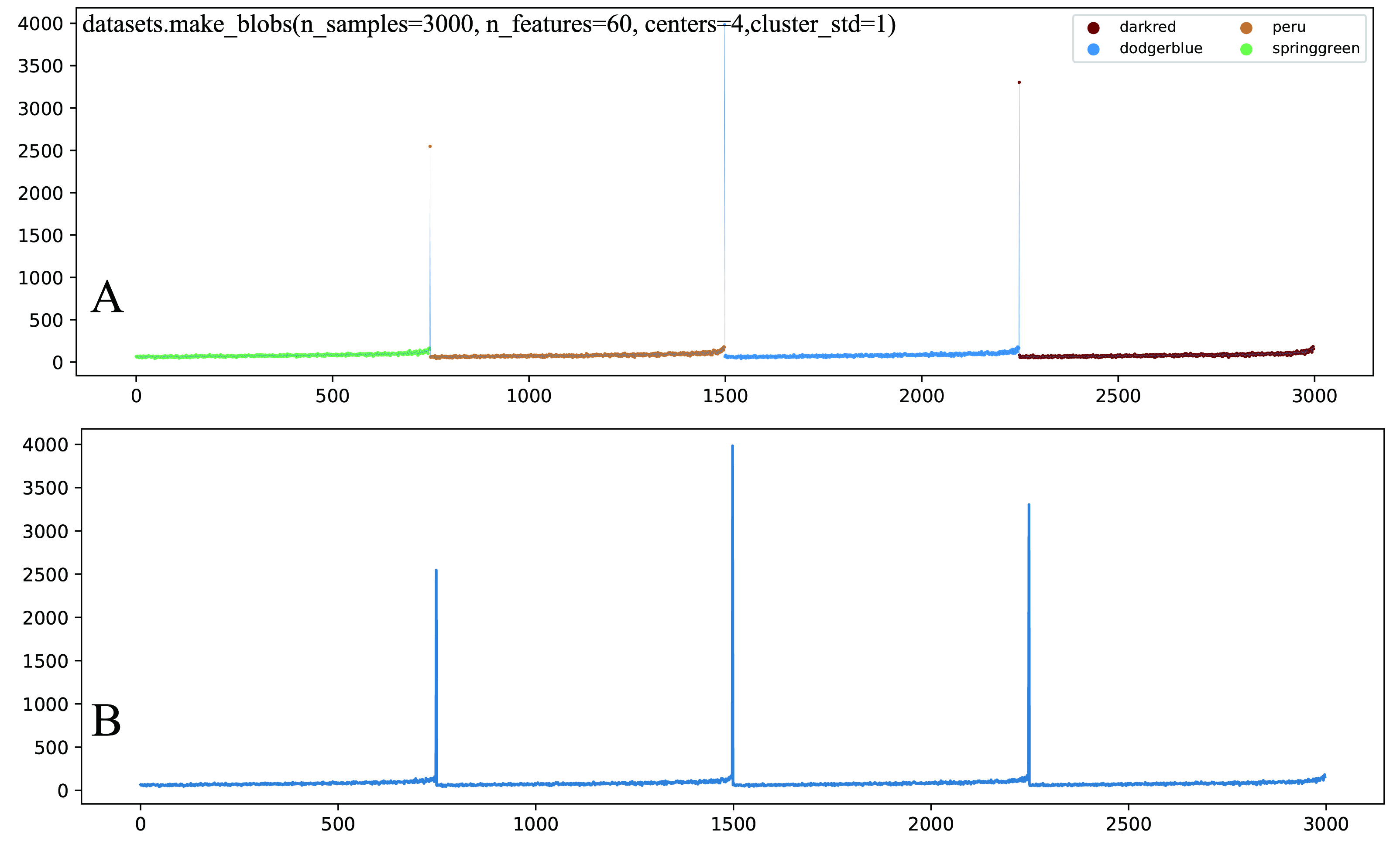}
\caption{DCSA applied to artificial data sets}
\label{fig:8}
\end{figure}

We Manually Generated a 4-class Datasets (MGD) with 3000 instances and 60 attributes. After applying DCSA to MGD, the results are shown in Figure \ref{fig:8}.

\begin{figure}[pos=H]
\centering
\includegraphics[width=0.8\textwidth]{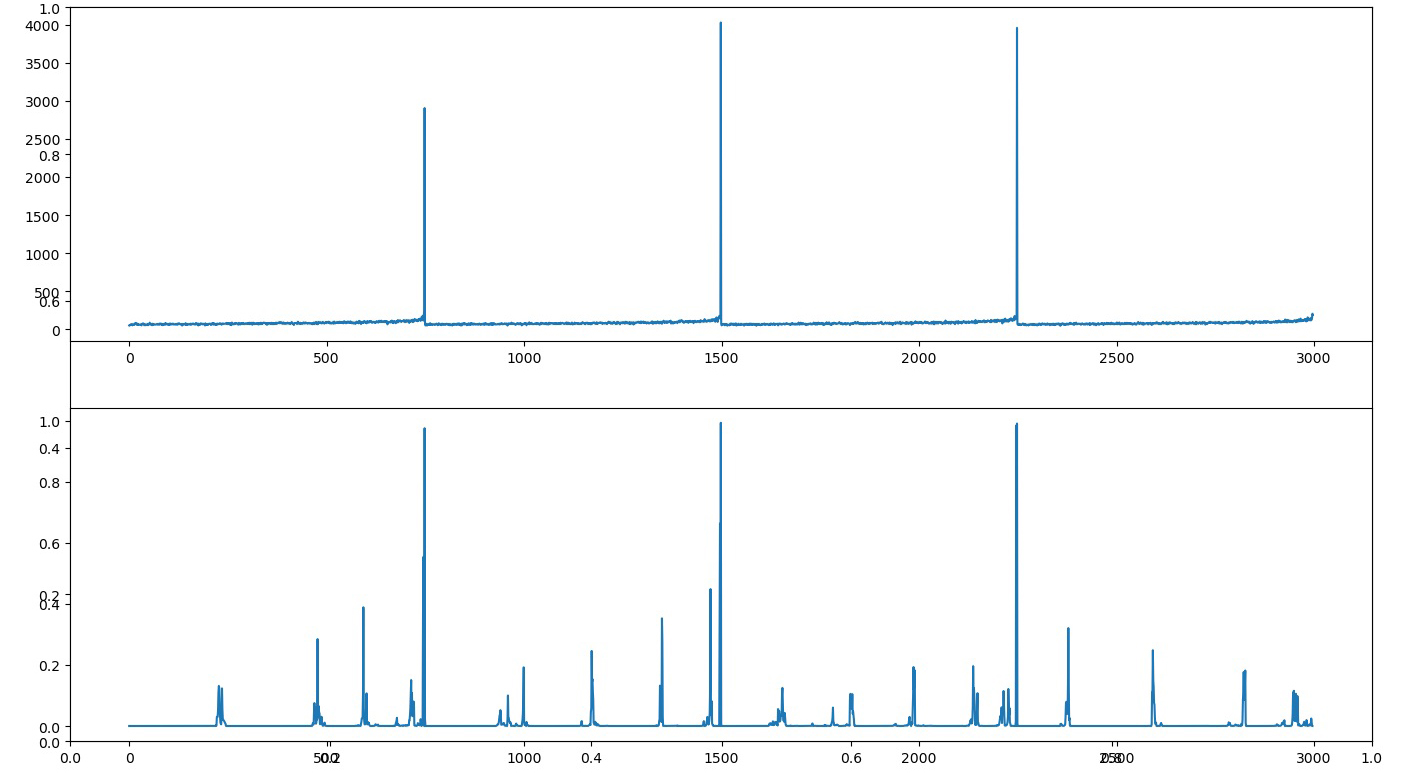}
\caption{Apply BCD to analyze artificial DataSet Sequence}
\label{fig:9}
\end{figure}

Figure \ref{fig:8}B is the sequence generated by DCSA, and its coloring results in Figure \ref{fig:8}A  shows that the DCSA divides the different classes into separate fragments, of which each fragment delivers an upward trend. It is indicated that this phenomenon is a feature of the DCSA sequence.Then we use BCD to analyze the sequence, and the result is shown in Figure \ref{fig:9}. The point where the probability is close to 1 is the correct cutoff point of the sequence.

The MGD dataset with isolated clustering structure are ideal, and rarely shown in real circumstances, and hence we need to further explore the performance of our algorithm with real-world dataset.
\subsection{Real-world data experiment}
In this section, DCSA is applied to the data set from UCI\citep{dua2017uci}\&KEEL\citep{alcala2009keel, alcala2011keel}, see the result in Table2 for the details.

\begin{table}[pos=H]
\centering\small
\begin{tabular}{lllllll}
\multicolumn{7}{l}{\small{\textbf{Table 2}}}\\
\multicolumn{7}{l}{\small{Real world dataset experiment}}\\\specialrule{0.05em}{3pt}{3pt}
$Datasets$&Records	&Attribute	&Cluster	&Distribution	&Area	&Source\\\specialrule{0.05em}{2pt}{2pt}
$1:digits$&1797&64&10&\tabincell{c}{178/182/177/183/181/\\182/181/179/174/180}&Graphic&UCI\\[7pt]
$2:wisconsin$&683&9&2&444/239&Medical&KEEL\\[7pt]
$3:ionosphere$&351&34&2&225/126&Engineering&KEEL\\[7pt]
$4:dermatology$&358&34&6&111/60/71/48/48/20&Medical&UCI\\[7pt]
$5:penbased$&10092&16&10&\tabincell{c}{1143/1143/1144/\\1055/1144/1055/1056\\/1142/1055/1055}&Graphic&KEEL\\[7pt]
$6:RNA-seq$&801&20531&4&135/141/300/146/78&Life&UCI\\[7pt]
\specialrule{0.05em}{2pt}{0pt}
\end{tabular}
\label{table:2}
\end{table}
\normalsize
In the previous section, The Spectral Clustering is the algorithm with the highest overall performance, and Kmeans is the most efficient clustering algorithm.  At the end of this section, will comparing DCSA with these two algorithms through AMI score.
\subsubsection{Digits dataset\citep{pedregosa2011scikit}}
\begin{figure}[pos=H]
\centering
\includegraphics[width=0.9\textwidth]{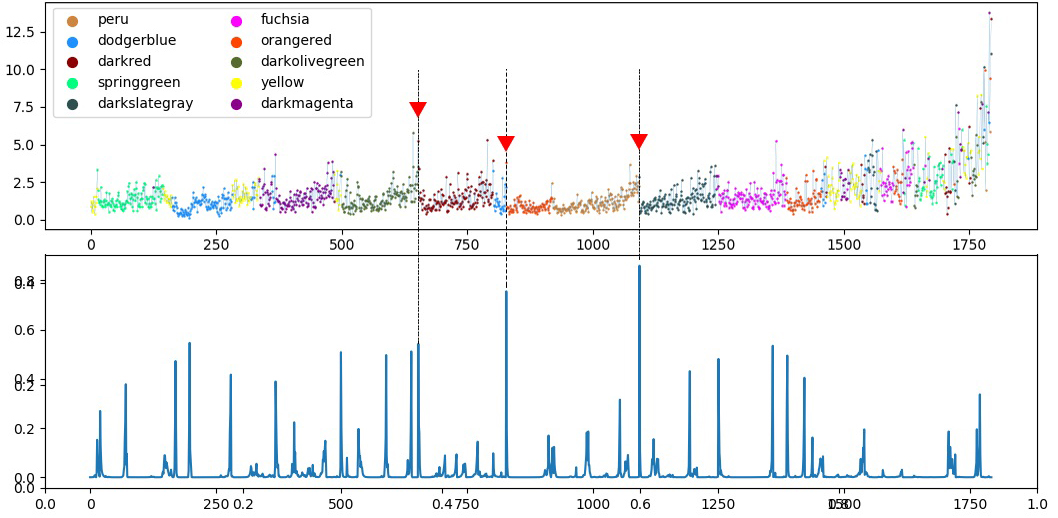}
\caption{Analysis digits DCSA sequence using BCD}
\label{fig:10}
\end{figure}

This high-dimensional data set has 1797 instances and 64-dimensional attributes.  It comes from the field of image recognition and is composed of handwritten Arabic numerals from 0 to 9.  As the first real-world data experiment, we used the method in section 3.3 to analyze the DCSA sequence.
\paragraph{BCD application}
\ 
\\\
\indent Figure \ref{fig:10} depicts the results of application effect of the DCSA sequence and BCD algorithm.

Except for the yellow and orange classes, most of other classes are concentrated on the same interval.  In Figure \ref{fig:10}  the red ``$\nabla $" masks the corresponding position, where the correct point found by BCD algorithm, and the probability of the three point is greater than 0.5.  However, no regularity can be observed at other points. It can be assumed that the BCD algorithm is relatively reliable at the position where the probability of the change point is close to 1.

\begin{figure}[pos=H]
\centering
\includegraphics[width=0.9\textwidth]{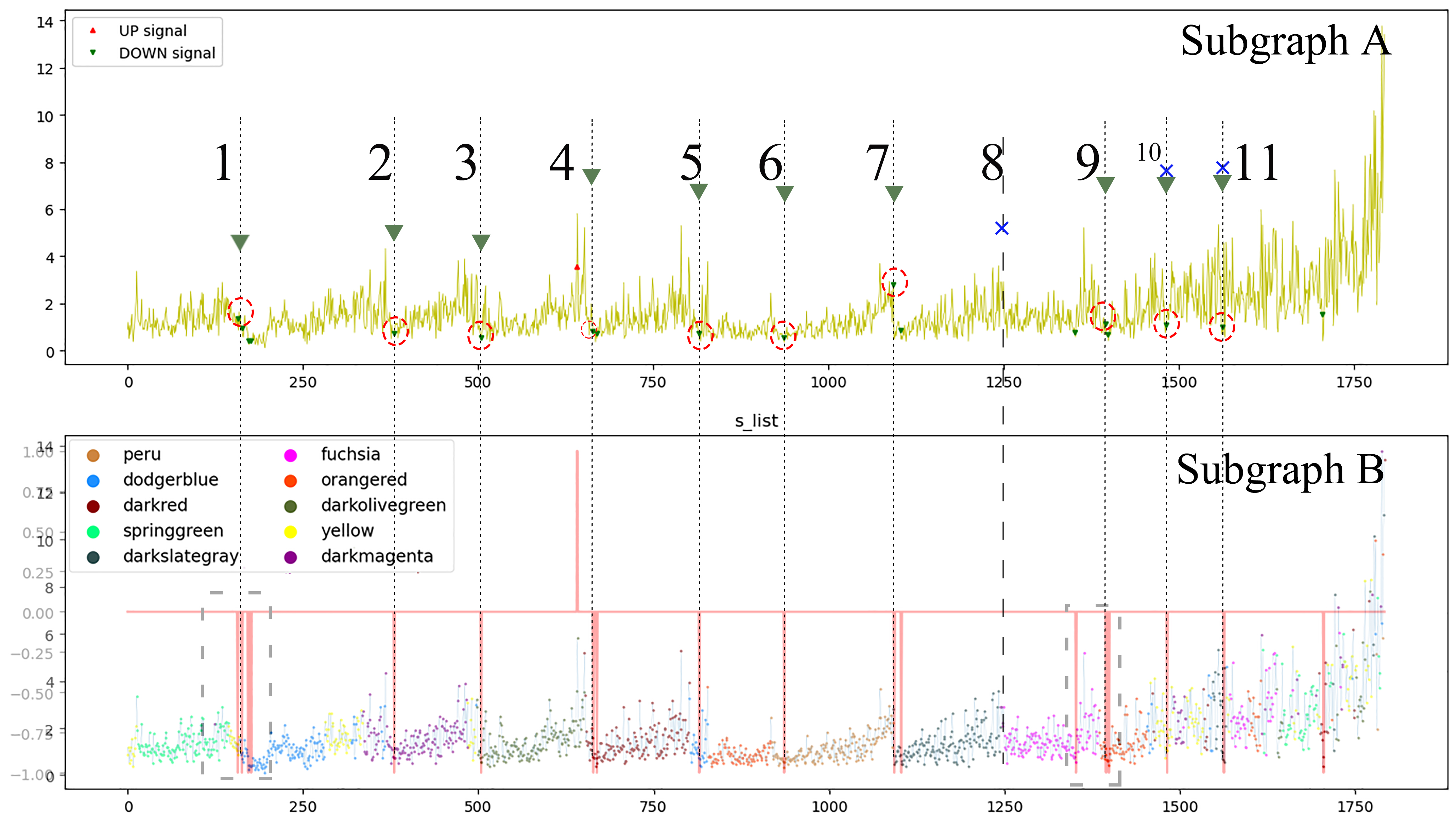}
\caption{CUSUM-A Analysis digits sequence}
\label{fig:11}
\end{figure}

\paragraph{CUSUM application}
\ 
\\\
\indent CUSUM requires two input parameters: Threshold and Accuracy.  The threshold determines the maximum cumulative change, and the accuracy determines the degree of perception of subtle changes. Figure \ref{fig:11} shows the results of the DCSA sequence coloring and the CUSUM-A algorithm application effect($T=19\And A=4$).
CUSUM-A Change Points are divided into ``upper CP" and ``lower CP".  The red line in Figure \ref{fig:11}B indicates the type and location of the change point.  The mark ``$\nabla $" illuminates the correct change point, and most of the ``lower CP" just fall At the best cutting point except the place where marked ``$ \times $" by author.  ``$\nabla $" with ``$ \times $" means wrong change point.

\begin{figure}[pos=H]
\centering
\includegraphics[width=0.95\textwidth]{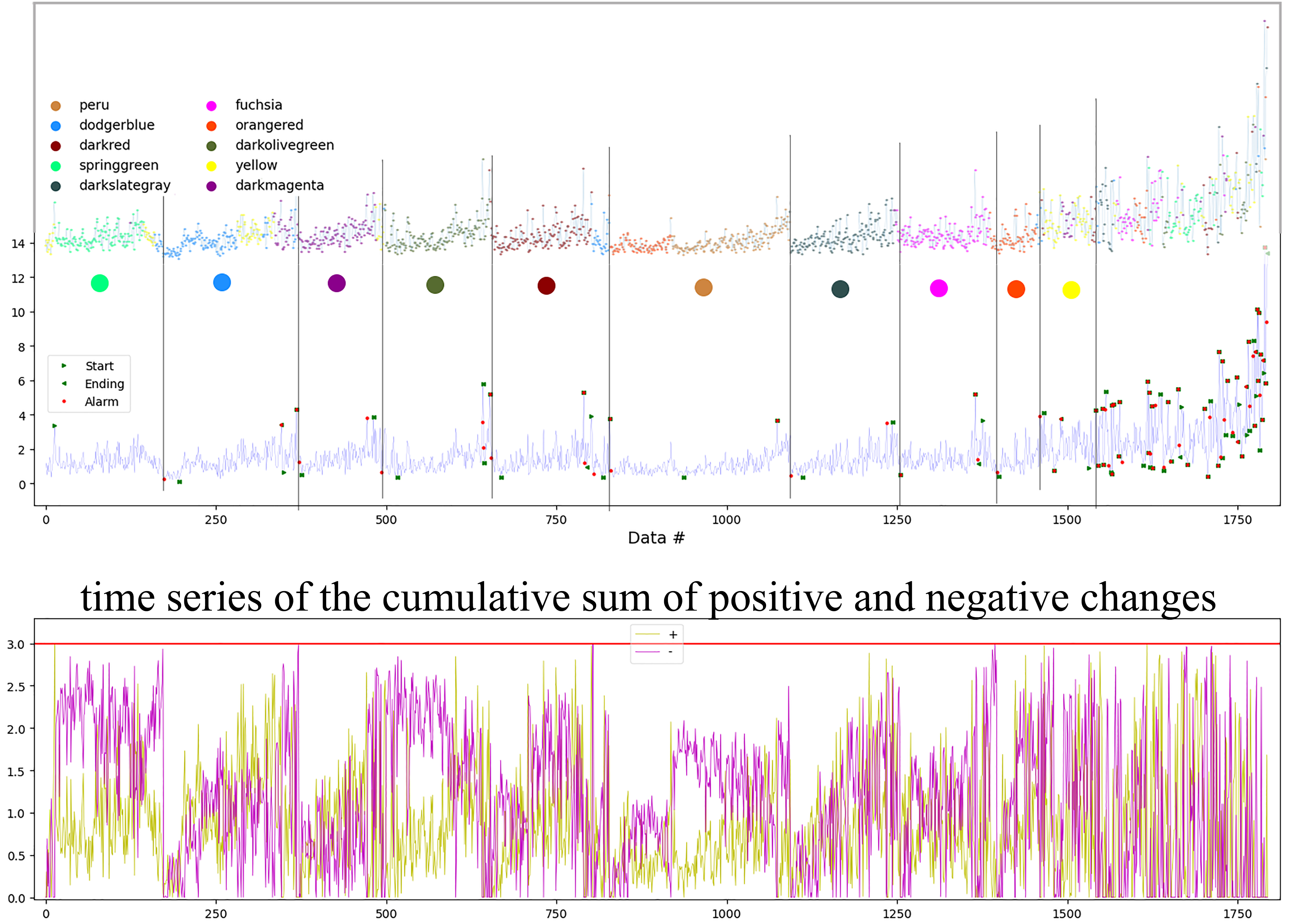}
\caption{CUSUM-B Analysis digits sequence}
\label{fig:12}
\end{figure}

CUSUM-A is the algorithm edited by the author of this paper.  On some specific data sets, CUSUM-B can get more detailed results, are shown in Figure \ref{fig:12}.
The red point (Alarm) in the figure is the change point.  It is shown that the location of the change point is exactly the classification point.

\begin{figure}[pos=H]
\centering
\includegraphics[width=0.9\textwidth]{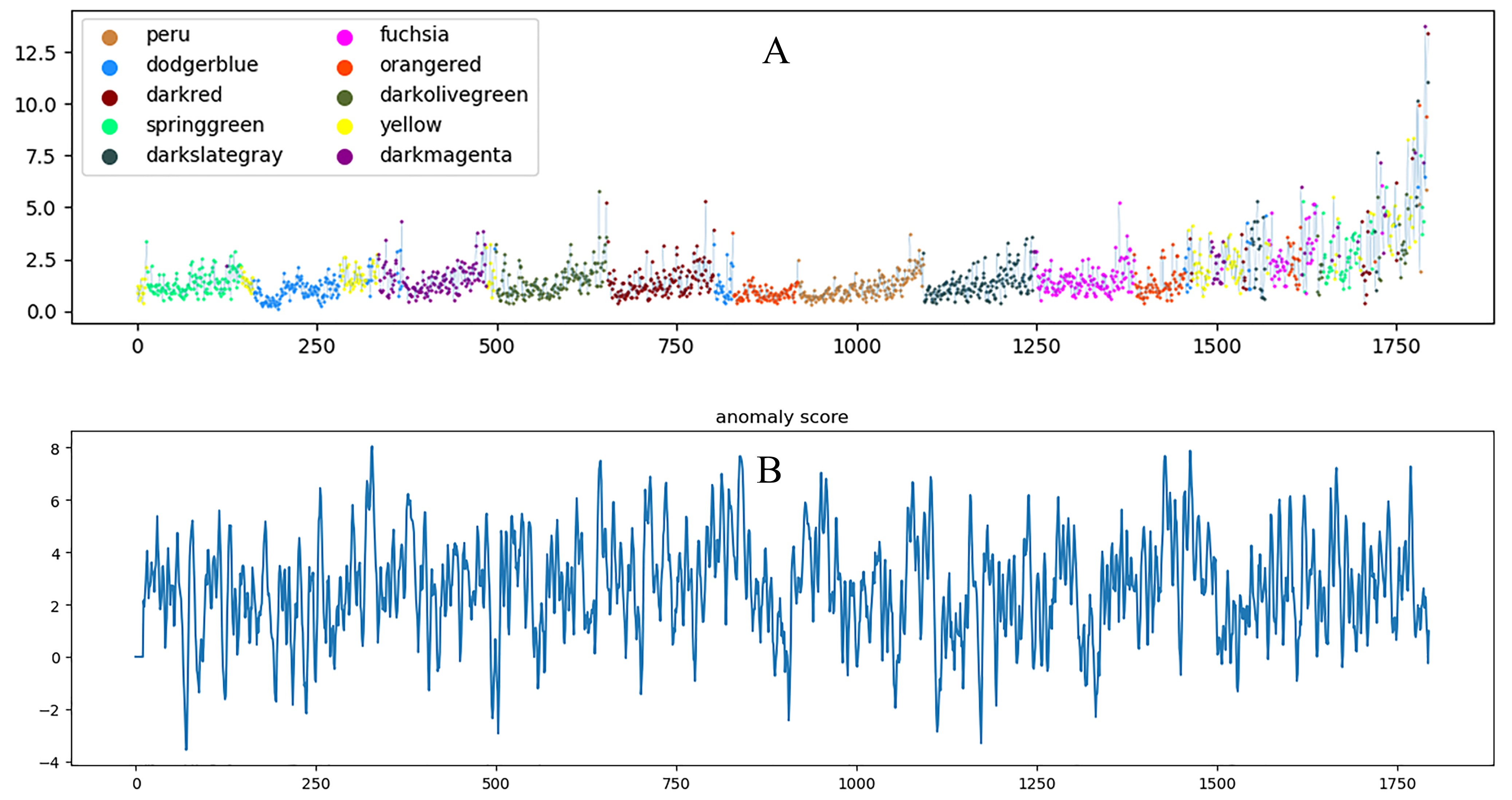}
\caption{ARIMA analysis digits sequence}
\label{fig:13}
\end{figure}

\noindent CUSUM-B change point filtering:

When too many change points are detected locally, the results need to be further calculated: set the distance threshold $d$ between the existing change point ${{\theta }_{i}}$ and the next change point ${{\theta }_{i+1}}$, When it is greater than $d$, the new change point is adopted, otherwise the change point is discarded.  Loop this process, and finally find all the change points of the division sequence.
\paragraph{ARIMA application}
\ 
\\\
\indent Figure \ref{fig:13} shows the calculation results of the ChangFinder function based on ARIMA.

\begin{figure}[pos=H]
\centering
\includegraphics[width=0.9\textwidth]{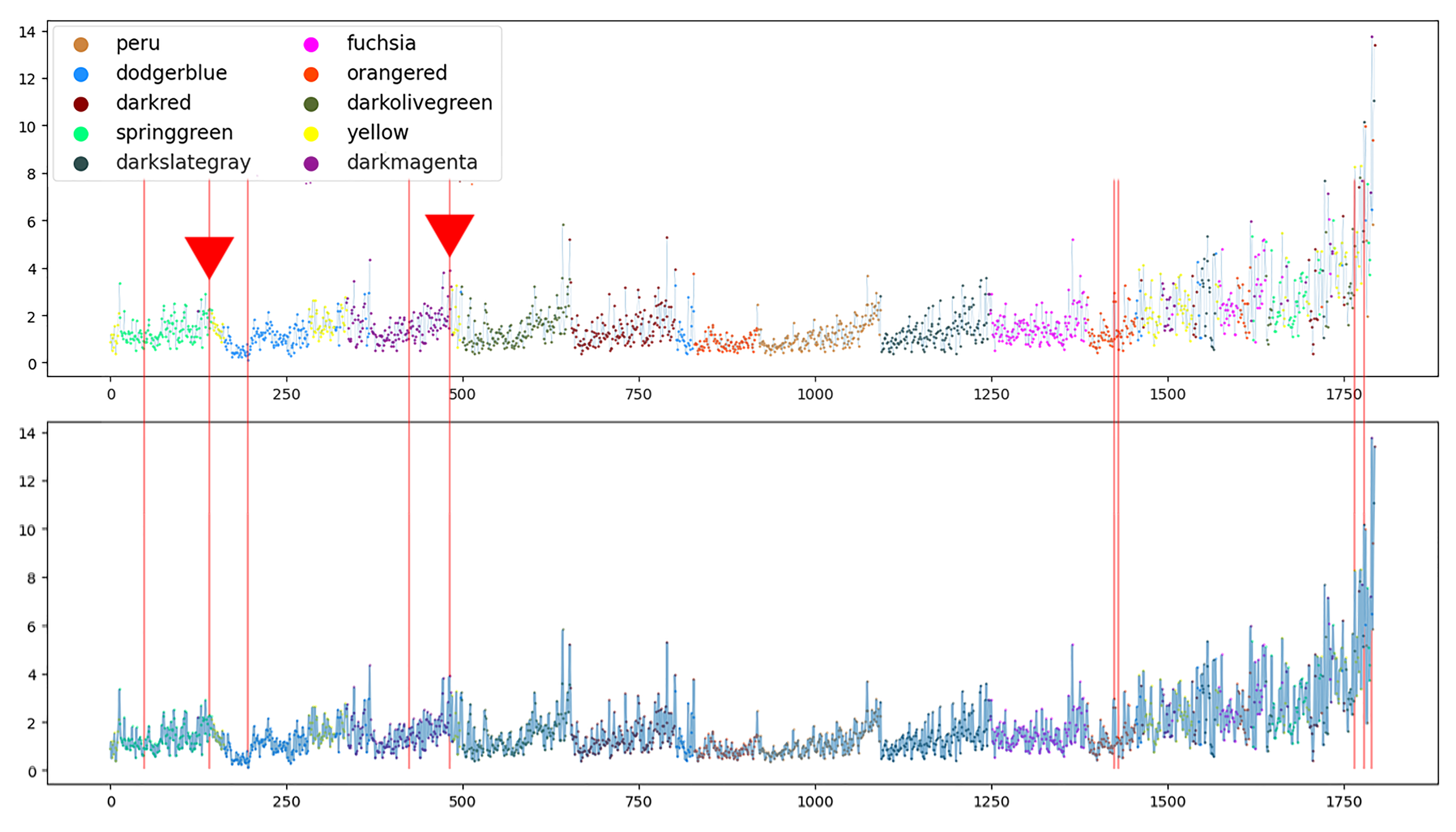}
\caption{Jenks analysis of digits generated sequence}
\label{fig:15}
\end{figure}

The analysis result of the ARIMA algorithm in Figure \ref{fig:13}B shows that the ARIMA is not suitable for the DCSA sequence analysis.
\paragraph{JNBK application}
\ 
\\\
\indent The positions of the change point are shown in Figure \ref{fig:15}.
Only some of the change points in Figure \ref{fig:15} are accurate, indicating the algorithm is not suitable for analysis the DCSA sequences.

The above experiments demonstrate that the CUSUM algorithm combined with the DCSA is effective on the digits data set.  Other data experiments also support this conclusion. Except Wisconsin, the other datasets only show the sequence coloring results.
\subsubsection{Wisconsin dataset}
\begin{figure}[pos=H]
\centering
\includegraphics[width=0.9\textwidth]{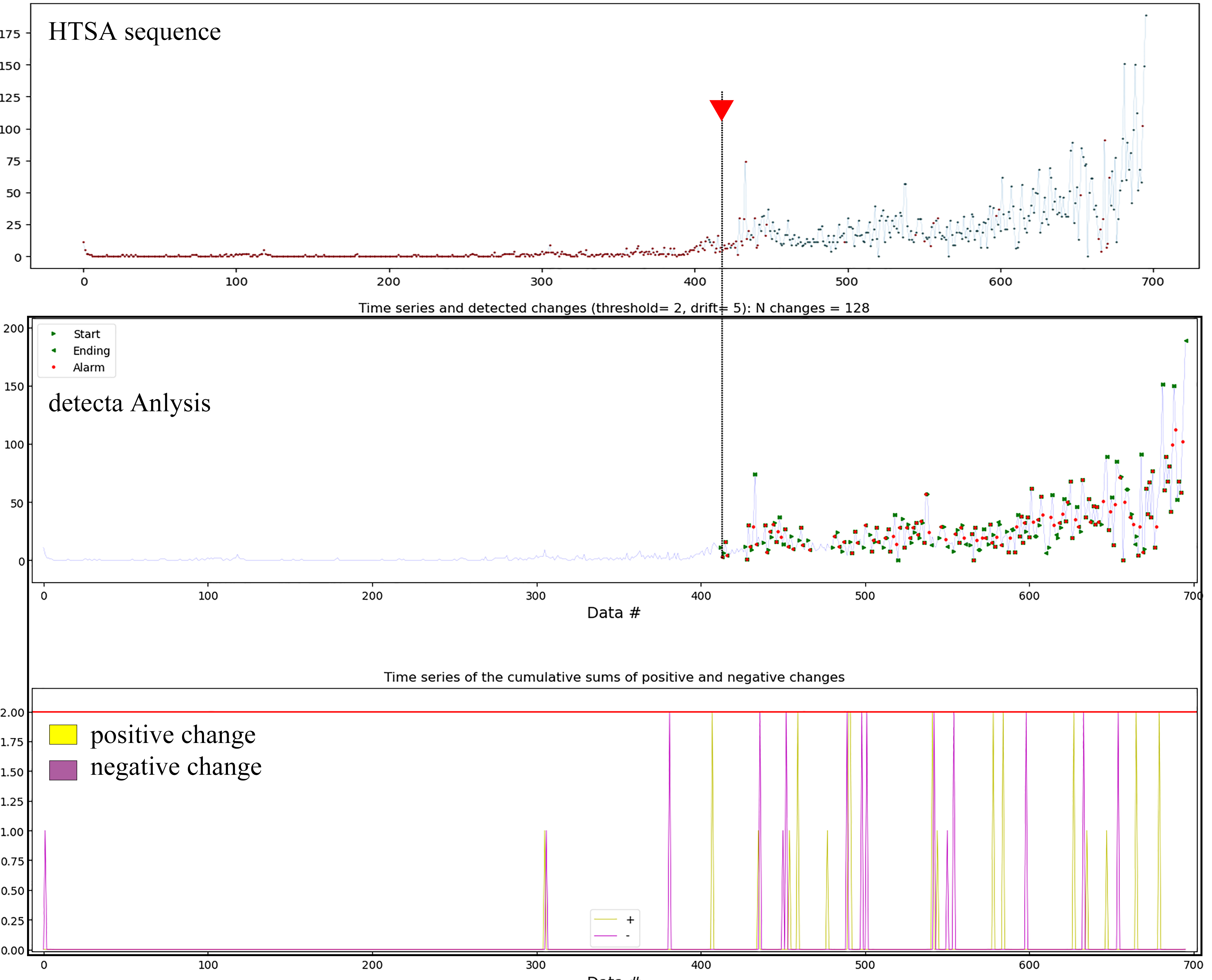}
\caption{CUSUM-B analysis of winsconsin generated sequence}
\label{fig:16}
\end{figure}

This medical data set contains cases of breast cancer surgery patients in a study conducted by the University of Wisconsin-Madison. It is an unbalanced low-dimensional data set. The goal of the experiment is to confirm whether the tumor detected by DCSA is benign or malignant. Figure \ref{fig:16} shows that the two categories are segmented correctly. The first one is relatively flat, while the latter shows an upward trend with fluctuations, and the connection between the two categories is relatively smooth. In this experiment, the CUSUM-B algorithm must be combined with the change point filtering in 4.2.1.2 to get the singal change point which is identified by the ``$\nabla $'' mark.
\subsubsection{Ionosphere datasets}
\begin{figure}[pos=H]
\centering
\includegraphics[width=0.9\textwidth]{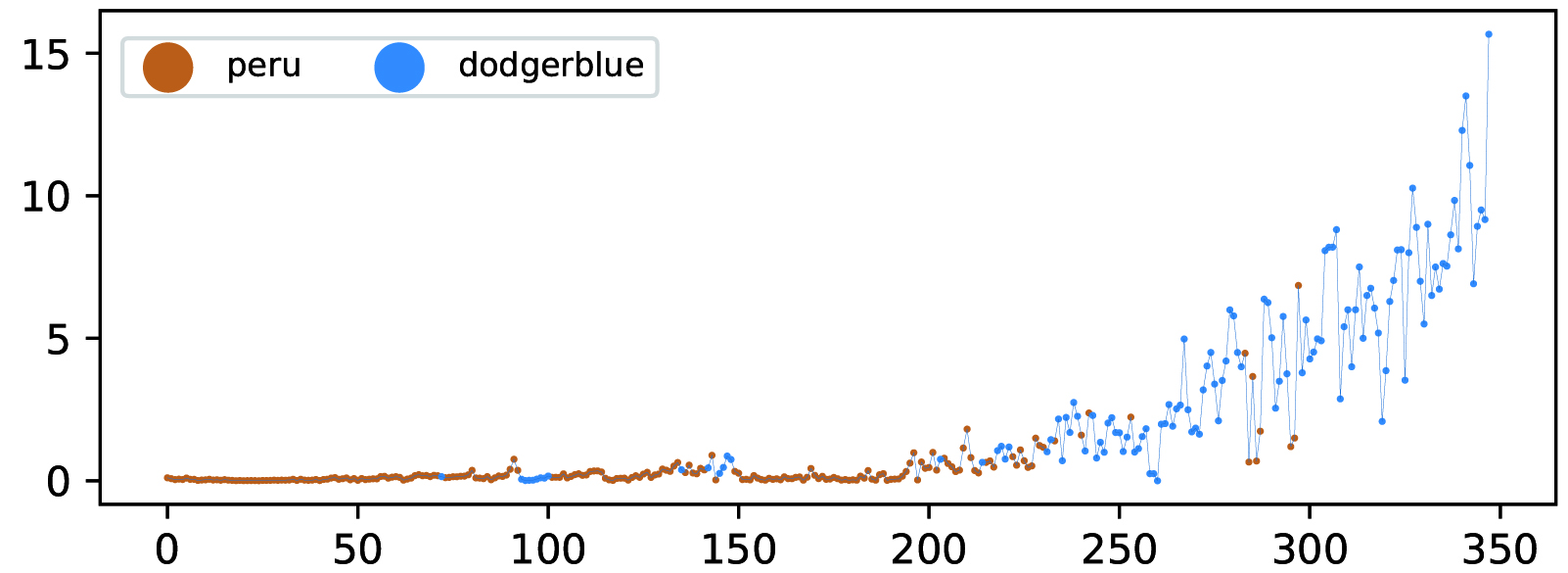}
\caption{Ionospheric dataset DCSA sequence}
\label{fig:18}
\end{figure}

This data set is the classification data of ionospheric radar echo. This data set focuses on the performance of DCSA under the two-class high-dimensional imbalance. Figure \ref{fig:18} shows a small amount of misclassification.
\subsubsection{Dermatology dataset}
\begin{figure}[pos=H]
\centering
\includegraphics[width=0.95\textwidth]{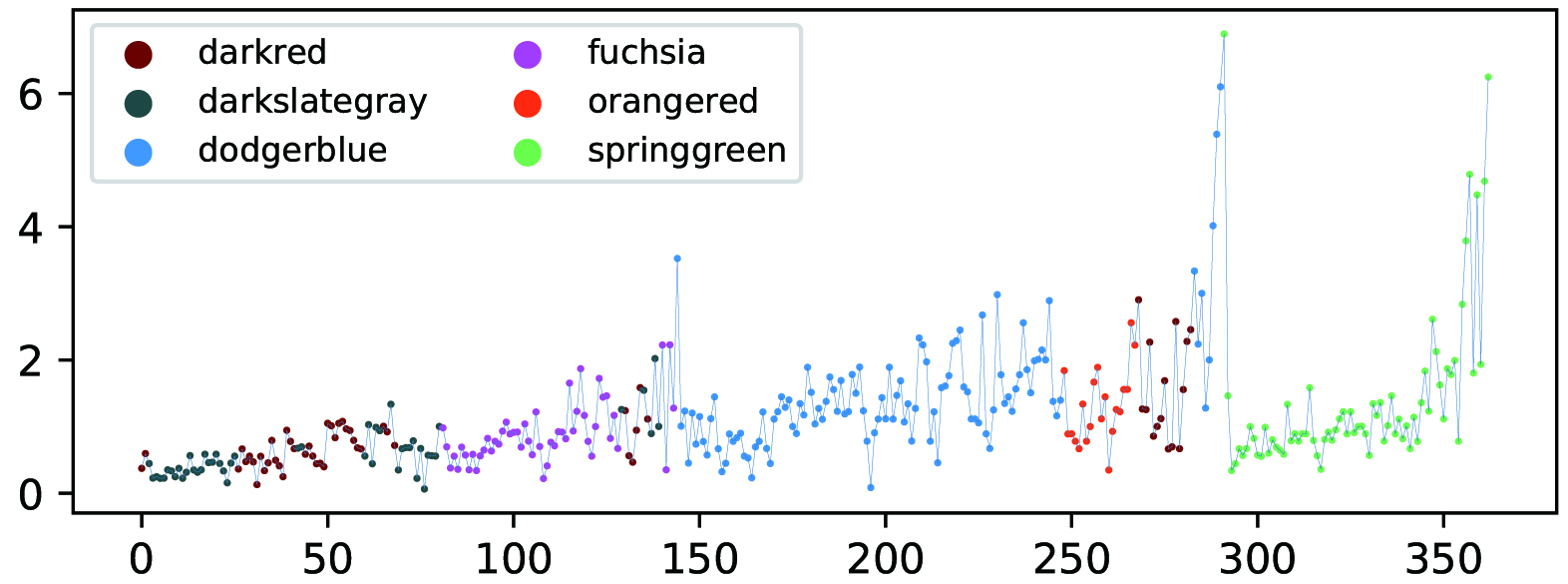}
\caption{dermatology dataset DCSA sequence}
\label{fig:19}
\end{figure}

This data set distinguishes the types of skin diseases based on characteristics. The data set is high-dimensional and unbalanced. The application effect of DCSA is shown in Figure \ref{fig:19}.
There are four categories in Figure \ref{fig:19} that are well divided. Darkred and darkslategray  have some confusion.
\subsubsection{Penbased dataset}
\begin{figure}[pos=H]
\centering
\includegraphics[width=0.9\textwidth]{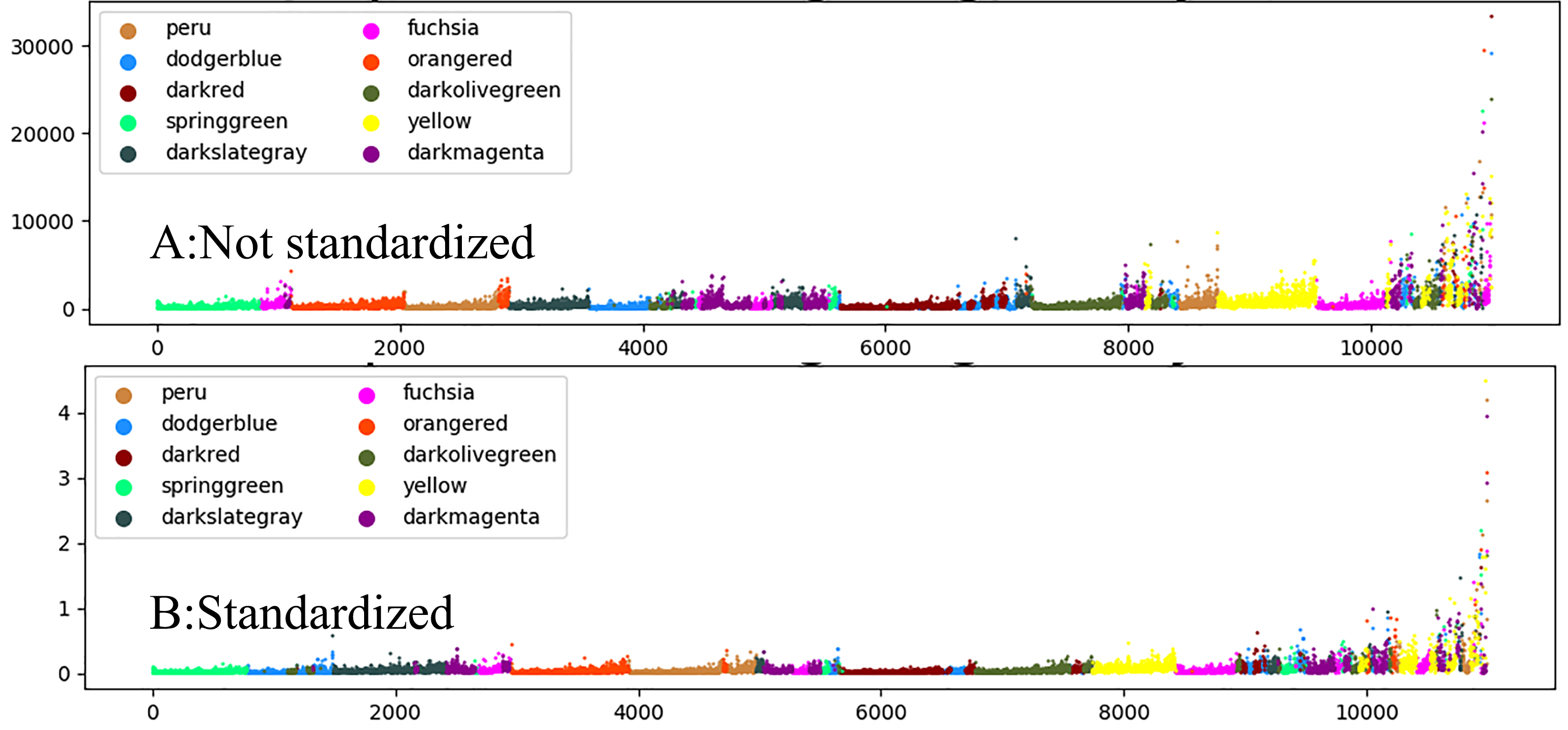}
\caption{Not standardization \& standardization DCSA sequence coloring by class}
\label{fig:20}
\end{figure}

Digit database of 250 samples from 44 writers is high-dimensional and balanced, and has the largest number of samples in this paper (10092). The results are shown in Figure \ref{fig:20}.
At the rear of the DCSA sequence of this dataset, various classes are mixed together. This should be caused by the overlapping of multiple classes in space.
\subsubsection{RNA-Seq dataset}
\begin{figure}[pos=H]
\centering
\includegraphics[width=0.9\textwidth]{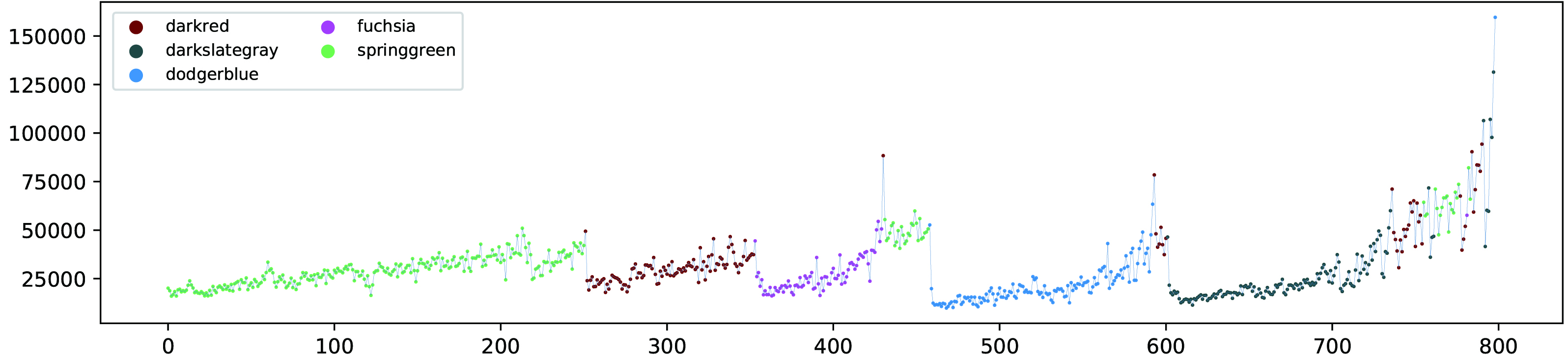}
\caption{DCSA sequence classification coloring of cancer RNA data}
\label{fig:21}
\end{figure}

This collection of data is part of the RNA-Seq (HiSeq) PANCAN data set, it is a random extraction of gene expressions of patients has different types of tumor: BRCA, KIRC, COAD, LUAD and PRAD. This data set is high-dimensional (the size reaches 20531), unbalanced and the size is larger than the number of samples. The algorithm performs better without standardizing the dataset. The classification results of DCSA algorithm is shown in Figure\ref{fig:21}.
Obviously, the DCSA application has achieved good results on this data set.
\subsubsection{Algorithm evaluation and comparison}
Table 3 compares the AMI scores of the three algorithms : Spectral Clustering, K-means, and DCSA.\\ In order to reproduce the results of the experiment, the parameters of spectral clustering are set as:\\assign\_labels="discretize",random\_state=0.

\begin{table}[pos=H]
\centering\small
\begin{tabular}{llll}
\multicolumn{4}{l}{\small{\textbf{Table 3}}}\\
\multicolumn{4}{l}{\small{Algorithm evaluation and comparison}}\\\specialrule{0.05em}{3pt}{3pt}
$Datasets$&DCSA&K-means &SpectralClustering\\\specialrule{0.05em}{2pt}{2pt}
$1:digits$&0.614068803&0.736937078&0.74029209\\[7pt]
$2:wisconsin$&0.644677955&0.731813974&0.000934103\\[7pt]
$3:ionosphere$&0.415083449&0.130530901&0.027340228\\[7pt]
$4:dermatology$&0.556677503&0.872912009&0.857759612\\[7pt]
$5:penbased$&0.534842473&0.665320729&N/A\\[7pt]
$6:RNA-seq$&0.687214727&0.977097745&-0.001459887\\[7pt]
$Mean\ score:$&0.575427485&0.685768739&0.270811024\\[7pt]
\specialrule{0.05em}{2pt}{0pt}
\end{tabular}
\label{table:3}
\end{table}
Spectral Clustering timed out on the Penbased dataset (more than 24 hours) and no results were output,When calculating the average score, we record it as 0.
\section{Conclusions}
Most data sets can be converted into numerical matrices after preprocessing. then the data has a feature space. The exploration of the feature space may reveal the latent patterns of data, DCSA is a new tools to do this. The experiment and visualization results on various types of data confirm the effectiveness of DCSA.

This version of DCSA application has experience requirements, because we need to choose a suitable sequence analysis algorithm, and these algorithms may need parameters such as thresholds. A stable sequence analysis method is an important prerequisite for the application of the DCSA.  Subsequent work will further study the combination of sequence analysis and DCSA.  In fact, if we don't have prior knowledge, we can identify patterns through the periodic upward trend in the DCSA sequence chart. If the data has latent classes and the distance between classes is larger, this trend will become obvious, as we saw on Figure \ref{fig:21}.

In addition, the DCSA algorithm also reveals some phenomena worthing further studing, such as why the same kind of data shows an upward trend in the DCSA sequence? In other words, why do experiments show that in most cases, when crossing classes, discovery factor(DF) go directly to the position with the highest density of the next class? The study of these phenomena may not only help us improve DCSA, but also give us a deeper understanding of the spatial distribution of data points.

\bibliographystyle{unsrt}
\bibliography{mybibfile}


\end{document}